\documentclass{article}

\usepackage[nonatbib,final]{neurips_2023}

\usepackage{amsmath,amsfonts,bm}

\def\eqref#1{equation~\ref{#1}}

\def\1{\bm{1}}

\DeclareMathAlphabet{\mathsfit}{\encodingdefault}{\sfdefault}{m}{sl}
\SetMathAlphabet{\mathsfit}{bold}{\encodingdefault}{\sfdefault}{bx}{n}

\usepackage[utf8]{inputenc} 
\usepackage[T1]{fontenc}    
\usepackage{hyperref}       
\usepackage{url}            
\usepackage{booktabs}       
\usepackage{amsfonts}       
\usepackage{nicefrac}       
\usepackage{microtype}      
\usepackage{xcolor}         
\usepackage{multirow}
\usepackage[normalem]{ulem}
\usepackage{graphicx}
\usepackage{authblk}
\usepackage{makecell}
\usepackage{wrapfig}
\usepackage{tabularray}

\title{SEED-X: Multimodal Models with Unified Multi-granularity Comprehension and Generation}

\begin{document}
\author{
\textbf{Yuying Ge$^{1\ast}$ \qquad Sijie Zhao$^{1\ast}$ \qquad Jinguo Zhu$^{1\ast}$ \qquad Yixiao Ge$^{1,2\dagger}$\\} \vspace{-5pt} 
\textbf{\qquad Kun Yi$^{2}$ \qquad Lin Song$^{1}$ \qquad Chen Li$^{2}$ \qquad Xiaohan Ding$^{1}$ \qquad Ying Shan$^{1,2}$}

$^{1}$Tencent AI Lab \qquad $^{2}$ARC Lab, Tencent PCG\\
}

\maketitle

\begin{figure}[h!]
\vspace{-40pt}
\centering
\makebox[\textwidth][c]{\includegraphics[width=0.96\linewidth]{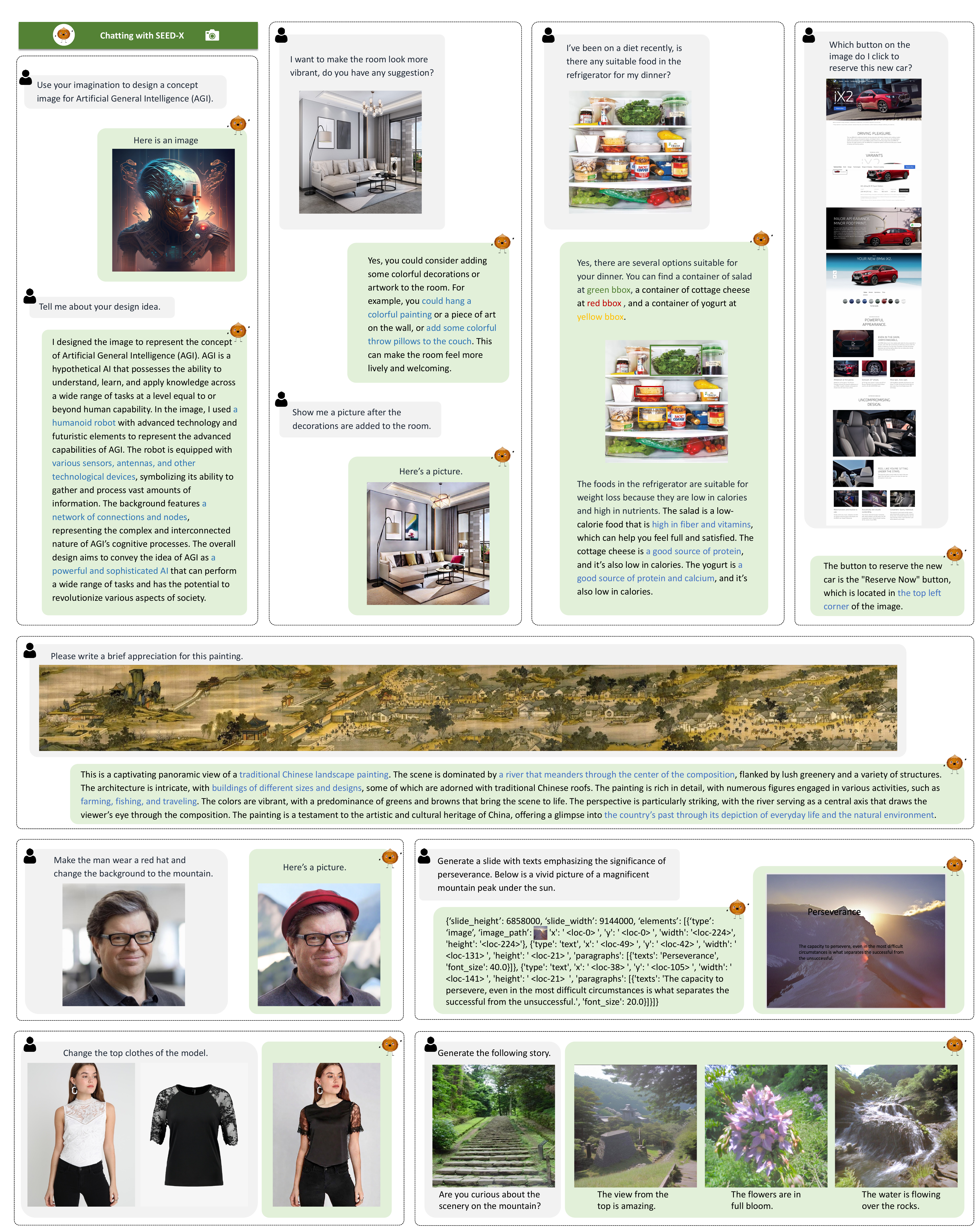}}%
\vspace{-5pt}
\caption{The introduced SEED-X, a unified and versatile foundation model, can serve as various multimodal AI assistants \textbf{in the real world} after different instruction tuning, capable of responding to a variety of user needs through unifying \textbf{ multi-granularity comprehension and generation}.}
\label{fig:teaser_example}
\vspace{-10pt}
\end{figure}

\begin{abstract}
The rapid evolution of multimodal foundation model has demonstrated significant progresses in vision-language understanding and generation, \textit{e.g.}, our previous work SEED-LLaMA. However, there remains a gap between its capability and the real-world applicability, primarily due to the model's limited capacity to effectively respond to various user instructions and interact with diverse visual data. In this work, we focus on bridging this gap through integrating two enhanced features: (1) comprehending images of arbitrary sizes and ratios, and (2) enabling multi-granularity image generation. We present a unified and versatile foundation model, namely, \textbf{SEED-X}, which is able to model multi-granularity visual semantics for comprehension and generation tasks. Besides the competitive results on public benchmarks, SEED-X demonstrates its effectiveness in handling real-world applications across various domains after instruction tuning. We hope that our work will inspire future research into what can be achieved by versatile multimodal foundation models in real-world applications. The models, codes, and datasets are released in \url{https://github.com/AILab-CVC/SEED-X}\footnote{This is the v2 version. We added benchmark results (without updating models) and ablation study.}.

\end{abstract}

\section{Introduction}
\label{sec:intro}

In recent years, Multimodal Large Language Models (MLLMs)~\cite{li2023blip2, zhu2023minigpt4, liu2023visual_llava, peng2023kosmos, bai2023qwen, liu2023llava1.5, zhang2023internlm, lin2023sphinx} have demonstrated exceptional capabilities in comprehending multimodal data through leveraging the strong generality of LLMs~\cite{touvron2023llama, brown2020language, chowdhery2022palm}. Some 
pioneering work~\cite{sun2023emu, yu2023scaling, ge2023planting, ge2023making, wu2023nextgpt, dong2023dreamllm, sun2023generative, zhu2023vl} further empower LLMs with the ability to generate images beyond texts. For example, our previous work SEED-LLaMA~\cite{ge2023making} can handle a variety of tasks and excel in academic benchmarks through unifying multimodal comprehension and generation. However, the accuracy and diversity of its generated content still fall short of real-world needs. In this work, we focus on bridging this gap through upgrading SEED-LLaMA with enhanced capabilities for real-world applications.

 \renewcommand{\thefootnote}{\fnsymbol{footnote}}
 		\footnotetext[1]{Equal Contribution. } 
   \footnotetext[2]{Correspondence to \texttt{yixiaoge@tencent.com}.}
 \footnotetext[3]{We sincerely acknowledge Tianheng Cheng (ARC Lab, Tencent PCG) for his support.}
 
Specifically, in order to make a multimodal foundation model applicable in real-world scenarios, we incorporate two enhanced features: (1) understanding images of arbitrary sizes and ratios, and (2) multi-granularity image generation, encompassing both high-level instructional image generation and low-level image manipulation tasks. These attributes can form the basis for a multimodal foundation model's effective application in an open-world context, since a multimodal foundation model has to accommodate various downstream tasks requiring different levels of visual semantics.

In this paper, we introduce SEED-X, a unified and versatile multimodal foundation model 
as a follow-up work of SEED-LLaMA, which seamlessly integrates the features mentioned above. It is important to emphasize that \textit{integrating all these characteristics into a single foundation model is by no means trivial}, as shown in Table~\ref{tab:comparison}, since none of the previous works support all of these features.

After different instruction tuning, SEED-X can function as various multimodal AI assistants in the real world, capable of addressing various user needs through generating proper texts and images as shown in Fig.~\ref{fig:teaser_example}. Specifically, our instruction-tuned models can act as an interactive designer, generating images while illustrating creative intent, offering modification suggestions and showcasing visualizations based on user's input images. Additionally, they can act as knowledgeable personal assistants, comprehending images of various sizes and providing relevant suggestions. Moreover, they can generate more diverse outputs, such as slide layouts for slide creation, and interleaved image-text content for storytelling. SEED-X signifies a notable advancement towards a versatile agent for users in the real world.

\begin{figure}[t]
  \vspace{-10pt}
	\centering
   \vspace{-10pt}
	\includegraphics[width=1.0\linewidth]{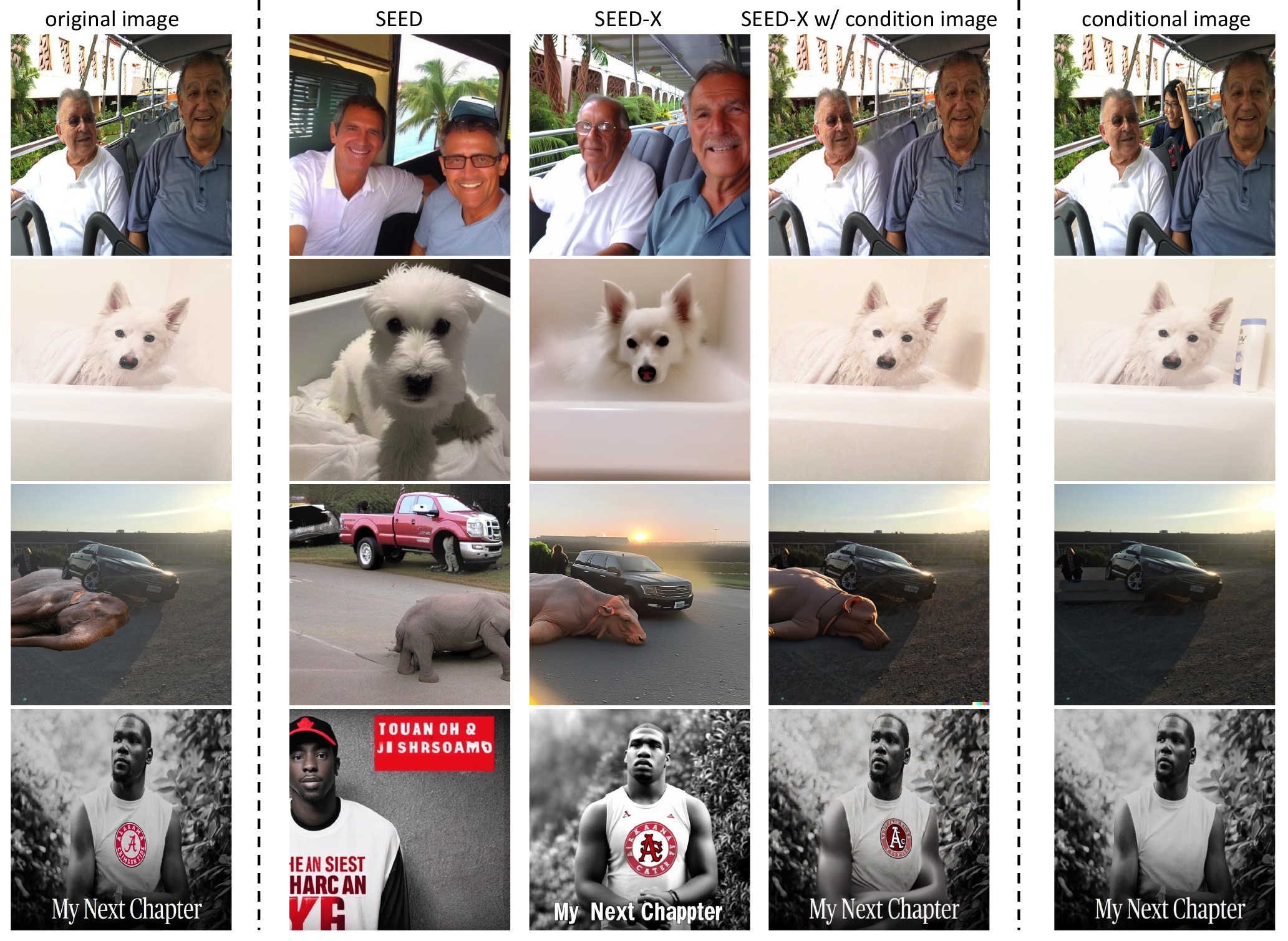}
\caption{The reconstruction results of our visual de-tokenizer. It can decode realistic images that are semantically aligned with the original images by taking the \textbf{ViT features} as inputs, and further recover fine-grained details by incorporating the \textbf{conditional images} as inputs. }
	\label{fig:tokenizer}
    \vspace{-5pt}
\end{figure}

To endow SEED-X with the aforementioned characteristics, our approach incorporates (1) a visual tokenizer to unify image comprehension and generation, where its multi-granularity de-tokenization phase facilitates image generation and high-precision image manipulation, and (2) an MLLM with dynamic resolution image encoding to enable the comprehension of images with arbitrary sizes and aspect ratios.
Specifically, we utilize a pre-trained ViT as the visual tokenizer and train a visual de-tokenizer to decode realistic images by taking the ViT features as input. 
To realize the retention of fine-grained details of the input image to satisfy image manipulation, we further fine-tune the visual de-tokenizer to take an extra condition image as input in the latent space (See Fig.~\ref{fig:tokenizer}). 
The ViT features \textbf{serve as a bridge} to decouple the training of the visual (de-)tokenizer and the MLLM.
The dynamic resolution image encoding divides an input image into sub-images and adds extrapolatable 2D positional embeddings to the ViT features of each sub-image, allowing the MLLM to scale to any image resolution. 
For image generation, a fixed number of learnable queries are fed into the MLLM, where the output hidden states are trained to reconstruct the ViT features of the target images. During inference, the image de-tokenizer can take both the output features from
the MLLM and the condition image provided by users as input, ensuring that the decoded image can
possess high-level semantics that meet the multimodal instructions and retain the low-level details.

We pre-train SEED-X on massive multimodal data, including image-caption pairs, grounded image-text data, interleaved image-text data, OCR data, and pure texts. We further apply multimodal instruction tuning to align SEED-X with human instructions across various domains, utilizing both existing datasets and newly collected datasets that cover image editing, text-rich, grounded and referencing QA, and slide generation tasks. The extensive evaluations on MLLM benchmarks demonstrate that our instruction-tuned model not only achieves competitive performance in multimodal comprehension, but also exhibits excellent instruction-following capabilities for image generation.

All the models, codes, and datasets are made publicly available. We hope our work can bring insights to the community about the potential of multimodal foundation models in real-world scenarios through unifying multi-granularity comprehension and generation.

\section{Related Work}
\label{sec:related}
With the rapid development of Multimodal Large Language Models (MLLM), recent studies have been working on unified MLLMs that are capable of \textbf{multimodal comprehension and generation} as shown in Tab.~\ref{tab:comparison}. Some work~\cite{ge2023making, ge2023planting, yu2023scaling, jin2023unified, lu2023unified, team2024chameleon, xie2024show, wu2024vila} utilize a discrete visual tokenizer to perform multimodal autoregression with a unified next-word-prediction objective or masked visual token prediction.
Some research efforts~\cite{sun2023emu, sun2023emu2, zhu2023vl} have delved into multimodal autoregression with continuous representations, where each image in the multimodal sequence is tokenized into embeddings via a visual encoder, and then interleaved with text tokens for autoregressive modeling. During inference, the regressed visual embeddings will be decoded into an image by a visual decoder. Additionally, some studies~\cite{dong2023dreamllm, wu2023nextgpt} enable image generation in a non-autoregressive manner through utilizing learnable queries to obtain visual representations from MLLMs, which are further fed into a image decoder to generate images. Mini-Gemini, generates text prompts using MLLMs and then leverages the existing SDXL~\cite{podell2023sdxl} to output images. 

Although these work have achieved competitive results on various academic benchmarks, such as VQA and text-to-image generation, the accuracy and diversity of
their generated content still fall short of real-world needs, since they do not meet the requirements of modeling multi-granularity visual semantics for comprehension and generation task. As shown in Tab.~\ref{tab:comparison}, we identify several significant characteristics essential for real-world applications including object detection and dynamic resolution image encoding for multi-granularity comprehension, as well as high-level instructional image generation and low-level image manipulation for multi-granularity image generation. Notably, \textbf{none of the previous works fully support all of these characteristics}.
In this work, we present SEED-X, a unified and versatile foundation model, which effectively incorporate the aforementioned characteristics for real-world applications.

\begin{table}[t]
\small
\centering
\caption{MLLMs that unify comprehension and generation and whether they support significant characteristics essential for real-world applications. ``Decoder Input'' denotes the inputs for image generation, where ``Features'' means continuous features, ``Token'' represents discrete tokens, ``Text'' implies text prompts.}
\label{tab:comparison}\vspace{5pt}
\resizebox{0.95\columnwidth}{!}{
\begin{tabular}{c|ccccccc}
\toprule
             & Date     & \begin{tabular}[c]{@{}c@{}}Decoder \\ Input\end{tabular} & \begin{tabular}[c]{@{}c@{}}Detec-\\ tion\end{tabular} & \begin{tabular}[c]{@{}c@{}}Dynamic \\ -Res Img\\ Input\end{tabular} & \begin{tabular}[c]{@{}c@{}}Image \\ Gen\end{tabular} & \begin{tabular}[c]{@{}c@{}}High-\\ precision \\ Editing\end{tabular} & \begin{tabular}[c]{@{}c@{}}Open-\\ source\end{tabular} \\\midrule
Emu        & 07/2023  & Feature                                                                                                          &$\times$                                                       &$\times$                                                           &$\checkmark$       &$\times$                                                                       &$\checkmark$                                                          \\
CM3Leon      & 07/ 2023 & Token                                                                                                         &$\times$                                                        &$\times$       &$\checkmark$                                                           &$\times$                                                                       &$\times$                                                         \\
SEED-OPT     & 07/ 2023 & Token                                                                                                         &$\times$                                                        &$\times$      &$\checkmark$                                                            &$\times$                                                                       &$\times$                                                         \\
LaVIT        & 09/2023  & Token                                                                                                           &$\times$                                                        &$\times$      &$\checkmark$                                                            &$\times$                                                                       &$\checkmark$                                                        \\
NExT-GPT     & 09/2023  & Feature                                                                                                       &$\times$                                                        &$\times$      &$\checkmark$                                                            &$\times$                                                                       &$\checkmark$                                                        \\
DreamLLM     & 09/2023  & Feature                                                                                                        &$\times$                                                        &$\times$         &$\checkmark$                                                         &$\times$                                                                       &$\times$                                                         \\
SEED-LLaMA & 10/2023  & Token                                                                                                         &$\times$                                                        &$\times$      &$\checkmark$                                                            &$\times$                                                                       &$\checkmark$                                                        \\
VL-GPT       & 12/2023  & Feature                                                                                                       &$\times$                                                        &$\times$        &$\checkmark$                                                          &$\times$                                                                       &$\times$                                                         \\
Gemini       & 12/2023  & Token                                                                                                               &$\times$                                                        & -             &$\checkmark$                                                   &$\times$                                                                       &$\times$                                                         \\
Emu2     & 12/2023  & Feature                                                                                                          &$\times$                                                        &$\times$    &$\checkmark$                                                              &$\times$                                                                       &$\checkmark$                                                        \\
Unified-IO 2 & 12/2023  & Token                                                                                                          &$\checkmark$                                                       &$\times$    &$\checkmark$                                                              &$\times$                                                                       & $\checkmark$                                                       \\
Mini-Gemini  & 03/2024  & Text                                                                                                           &$\times$                                                        &$\times$      &$\checkmark$                                                            &$\times$                                                                       &$\checkmark$                                                        \\
\textbf{SEED-X}     & 04/2024  & {Feature}                                                                                                   &{$\checkmark$}                                                       &{$\checkmark$}      &$\checkmark$                                                           &{$\checkmark$}                                                                      &{$\checkmark$}      \\               \bottomrule
\end{tabular}}
\vspace{-10pt}
\end{table}

\section{Method}
\subsection{Visual Tokenization and De-tokenization}
\begin{figure}[t]
	\centering
	\includegraphics[width=1.0\linewidth]{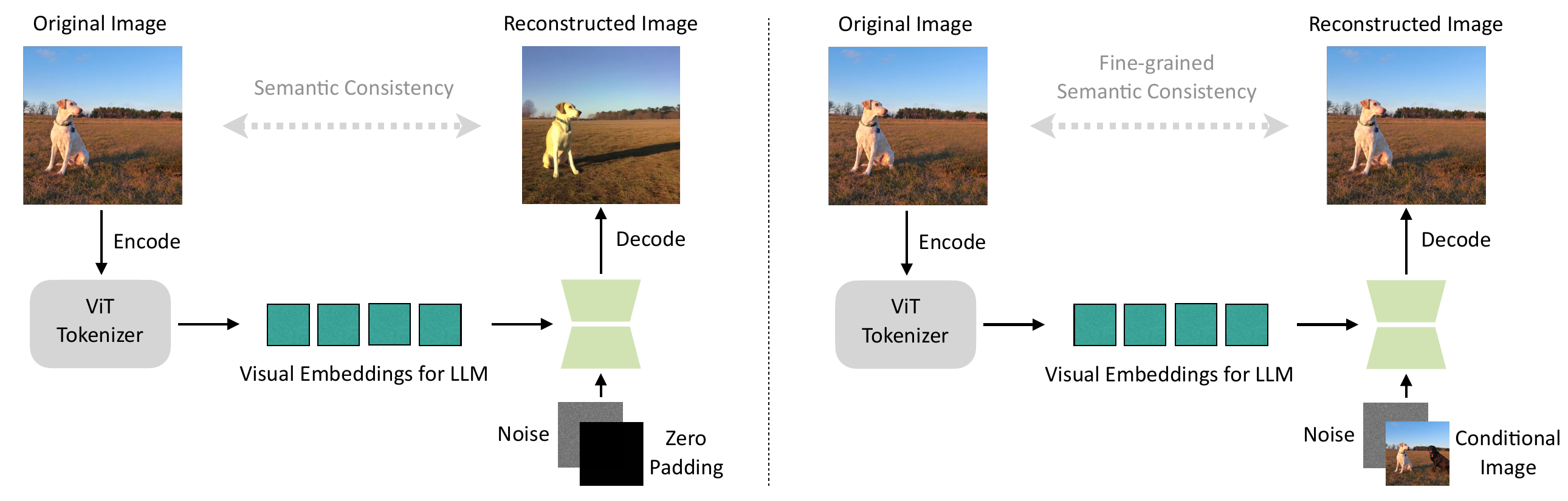}
\caption{Overview of visual tokenization and de-tokenization in SEED-X. In the first stage (left), we pre-train a visual de-tokenizer, which can decode semantically consistent images by taking the features of a pre-trained ViT as inputs. In the second stage (right), we fine-tune the visual de-tokenizer through concatenating the latent features of a conditional image with the noise to recover the fine-grained details of the original image.}
	\label{fig:sd_method}
\end{figure}

In SEED-X, we adopt a visual tokenizer to unify image comprehension and generation, and pre-train a multi-granularity
de-tokenizer to facilitate image generation and high-precision image manipulation in a two-stage manner. In the first stage, as shown in Fig.~\ref{fig:sd_method} (left), we utilize a pre-trained ViT as the visual tokenizer and pre-train a visual de-tokenizer to decode realistic images by taking the features of the ViT as inputs in the first stage. Specifically, $N$ visual embeddings from the ViT tokenizer ($N=64$ after average pooling) are fed into a learnable module as the inputs of the U-Net of the pre-trained SD-XL~\cite{podell2023sdxl} (replacing the original text features). The learnable module consists of four cross-attention layers to connect the visual tokenizer and the U-Net. We optimize the parameters of the learnable module and keys and values within the U-Net on the images from JourneyDB \cite{sun2024journeydb}, LAION-Aesthetics \cite{laion_aesthetics}, Unsplash \cite{unsplash}, and LAION-COCO \cite{laioncoco}. As shown in Fig.~\ref{fig:tokenizer}, compared with SEED~\cite{ge2023making}, our visual de-tokenizer can decode images that are more semantically aligned with the original images by taking the ViT features as inputs.

In the second stage, as shown in Fig.~\ref{fig:sd_method} (right), we further fine-tune the visual de-tokenizer to take an extra condition image as inputs for the retention of low-level details. Specifically, we follow InstructPix2Pix~\cite{brooks2023instructpix2pix} to encode the condition image into the latent space via the VAE encoder, and concatenate them with the noisy latent as the input of U-Net. The channel number of the U-Net convolutional layer is expanded from 4 to 8, and all parameters of U-Net are optimized. We fine-tune the visual de-tokenizer on MagicBrush~\cite{zhang2023magicbrush} and in-house image editing data, as well as the pure images in the first stage, where the conditional inputs are set to zeros. As shown in Fig.~\ref{fig:tokenizer}, by incorporating the condition image as an additional input besides the high-level image features, our visual de-tokenizer can recover the fine-grained details of the original image.

\subsection{Dynamic Resolution Image Encoding}
Current MLLMs require to resize the input images to a pre-defined resolution (typically a square size), which corresponds to the training resolution of the vision encoder, which can result in the loss of  fine-grained information. In this work, we propose dynamic resolution image encoding to enable the processing of images with arbitrary sizes and aspect ratios by dividing the image into a grid comprising of sub-images. Specifically, for the visual encoder with the training resolution $H_t \times W_t $, we first up-sample the input image with the size $H\times W$ to the size of $\{N_h*H_t\} \times \{N_w*W_t\}$. The grid size $N_h\times N_w$, are determined by
\begin{equation}
\small
\begin{aligned}
\min & \quad N_h * N_w, \\
\text { s.t. } & H \leq N_h * H_t \quad \text{and} \quad  W \leq N_w *  W_t.
\end{aligned}
\end{equation}
We also resize the original image to the size of $H_t\times W_t$ to provide global visual context. All sub-images and the resized global image are fed into the visual encoder to obtain the features, which are concatenated as the input of the LLM.

To enable the LLM to be aware of the positional information of each sub-image within the original image, we add extrapolatable 2D positional embeddings to the visual features of each sub-image. Specifically, for a sub-image with a normalized center location $(x_c, y_c)$ in the grid, where $0.0 \textless x_c, y_c \textless 1.0$, its learnable positional embedding $p$ is computed:
\begin{equation}
\small
\begin{aligned}
    p = & x_c * l + (1-x_c) *r + y_c * t + (1-y_c) *b.
\end{aligned}
\end{equation}
$l$,  $r$,  $t$, and $b$ represent four learnable position embeddings indicating left, right, top and bottom respectively. Consequently, our visual encoder can handle inputs with any arbitrary sizes and aspect ratios, even if the image resolution was not encountered during training.

\subsection{Multimodal Pre-training and Instruction Tuning}
\subsubsection{Training Stage I: Multimodal Pre-training}
\begin{figure}[t]
	\centering
	\includegraphics[width=1.0\linewidth]{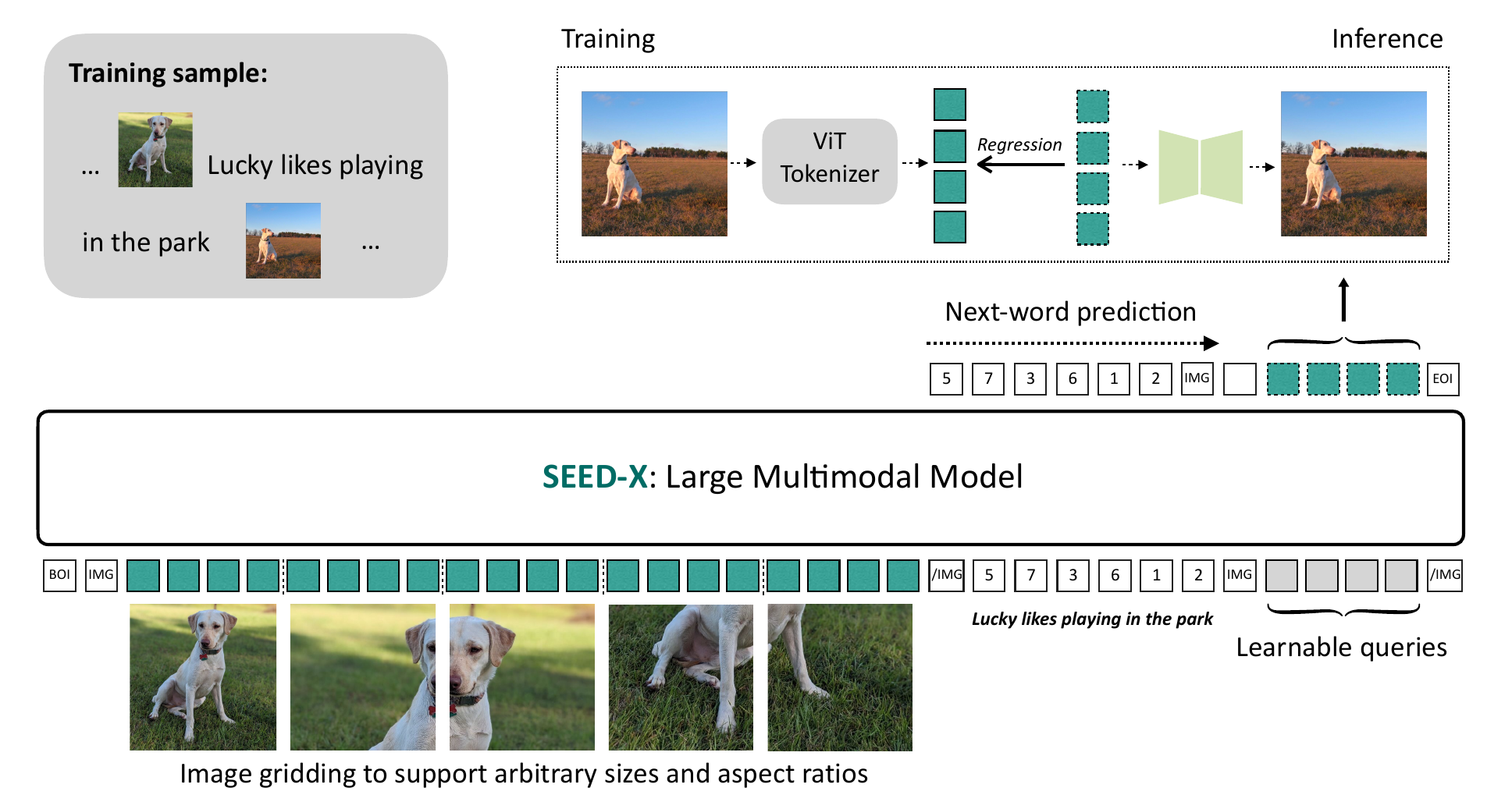}
\caption{Overview of SEED-X for multimodal pre-training. Each image is divided into sub-images to support arbitrary sizes and aspect ratios, and their ViT features along with text tokens are fed into an LLM to perform next-word prediction and image feature regression between the output hidden states of the learnable queries and ViT features. During inference, the regressed image features are fed into the visual de-tokenizer to decode images.}
	\label{fig:llm_method}
\end{figure}

As shown in Fig.~\ref{fig:llm_method}, SEED-X adopts next-word prediction and image feature regression training objectives on interleaved visual and textual data. Specifically, we perform dynamic resolution encoding of each image in the multimodal sequence, and their features along with text tokens are fed into the pretrained LLM. In order to equip the model with detection and referencing abilities, we add 224 bbox tokens, designated for representing bounding box coordinates, represented by \textless box\_start\textgreater{} \textless loc-x\_center\textgreater{} \textless loc-y\_center\textgreater{} \textless loc-width\textgreater{} \textless loc-height\textgreater{} \textless box\_end\textgreater{} with special tokens at the beginning and end of the bounding box. The text and added bbox tokens are trained through predicting the next token with cross-entropy loss.

We employ $N$ learnable queries ($N=64$ to align with the visual de-tokenizer) to obtain the output visual representations from the LLM, which are trained to reconstruct the features of the pre-trained ViT tokenizer with a Mean Squared Error (MSE) loss. We add two special tokens `\verb|<IMG>|' and `\verb|</IMG>|' to represent the beginning and the end of the query embeddings, and the `\verb|<IMG>|' is trained to predict where an image emerges. In doing so, we utilize the pre-trained ViT tokenizer as a \textbf{bridge} to decouple the training of a visual de-tokenizer and the MLLM for image generation. During inference, the regressed visual representations from SEED-X are fed into the visual de-tokenizer to decode realistic images.

We pre-train SEED-X initialized from Llama2-chat-13B using LoRA on massive multimodal data, including image-captions pairs, grounded image-texts, interleaved image-text data, OCR data and pure texts. We perform pre-training with 48 H800-80G GPUs (10 days) on a total of 158M samples. See Appendix.~\ref{sec:appendix_data} and Appendix.~\ref{sec:appendix_im} for more details.

\begin{table}[ht]
    \centering
    \setlength{\tabcolsep}{1.05pt}
    \renewcommand{\arraystretch}{1.2}
    \scriptsize
    \caption{\textbf{Comparison on multimodal understanding benchmarks}. ``Und.'' and ``Gen.'' denote ``understanding'' and ``generation'', respectively. }
    \begin{tabular}{llcccccccc}
        \toprule
        \textbf{Type} & \textbf{Model}  & \textbf{POPE$ \uparrow$} & \textbf{MME-P$ \uparrow$} & \textbf{MMB$ \uparrow$} & \textbf{SEED$_{\text{(img)}}$$ \uparrow$} & \textbf{VQAv2$_{\text{(test)}}$$\uparrow$} & \textbf{GQA$ \uparrow$} & \textbf{MMMU$ \uparrow$} & \textbf{MM-Vet$ \uparrow$} \\
        \midrule
        \textit{Und. Only} & 
        LLaVA-v$1.5$-Phi-$1.5$~\cite{xie2024show}  & $84.1$ & $1128.0$ & - & - & $75.3$ & $56.5$ & $30.7$ & - \\
        & MobileVLM~\cite{chu2023mobilevlm}  & $84.5$ & $1196.2$ & $53.2$ & - & - & $56.1$ & - & -\\
        & MobileVLM-V2~\cite{chu2024mobilevlm2}  & $84.3$ & $1302.8$ & $57.7$ & - & - & $59.3$ & - & -\\
        & LLaVA-Phi~\cite{zhu2024llava}  & $85.0$ & $1335.1$ & $59.8$ & - & $71.4$ & - & - & $28.9$\\
        & LLaVA~\cite{liu2024visual} & $76.3$ & $809.6$ & $38.7$ & $33.5$ & - & - & - & $25.5$ \\
        & LLaVA-v$1.5$~\cite{liu2024improved} &$\mathbf{85.9}$ & $\mathbf{1510.7}$ & $64.3$ & $58.6$ & $\mathbf{78.5}$ & $\mathbf{62.0}$ & $35.4$ & $31.1$ \\
        & InstructBLIP~\cite{instructblip}  & - & - & $36.0$ & $53.4$ & - & $49.2$ & - & $26.2$ \\
        & IDEFICS-$9$B~\cite{laurencon2023introducing} & - & - & $48.2$ & - & $50.9$ & $38.4$ & - & - \\
        & Qwen-VL-Chat~\cite{bai2023qwen}  & - & $1487.5$ & $60.6$ & $58.2$ & $78.2$ & $57.5$ & - & - \\
   
        \midrule
        \textit{Und. and Gen.} 
        & DreamLLM~\cite{dong2023dreamllm}  & - & - & - & - & $72.9$ & - & - & $36.6$ \\
        & LaVIT~\cite{jin2023unified}  & - & - & - & - & $66.0$ & $46.8$ & - & - \\
        & Emu~\cite{sun2023generative}  & - & - & - & - & $52.0$ & - & - & - \\
        & NExT-GPT~\cite{wu2023next}  & - & - & - & - & $66.7$ & - & - & - \\
        & Gemini-Nano-1~\cite{team2023gemini}  & - & - & - & - & $62.7$ & - & $26.3$ & - \\
        & LWM~\cite{liu2024world}  & $75.2$ & - & - & - & $55.8$ & $44.8$ & - & $9.6$ \\
        & \textbf{SEED-X}  & $84.1$ & $1457.0$ & $\mathbf{70.1}$ & $\mathbf{66.5}$ & $71.2$ & $49.1$ & $\mathbf{35.6}$ & $\mathbf{43.0}$ \\
        \bottomrule
    \end{tabular}
    \label{tab:result_understanding}
\end{table}

\subsubsection{Training Stage II: Multimodal Instruction Tuning}
We perform multimodal instruction tuning through fine-tuning SEED-X using a LoRA module with both public datasets and in-house data covering image editing, text-rich, grounded and referencing QA, and slide generation tasks. The details of datasets can be found in Appendix.~\ref{sec:appendix_data}. 
We fine-tune SEED-X with conversational and image generation data to yield a general instruction-tuned model SEED-X-I, which can follow multimodal instructions and make responses with images, texts and bounding boxes in multi-turn conversation. We further fine-tune the foundation model SEED-X on specialized datasets, resulting in a series of instruction-tuned models tailored for specific tasks, including SEED-X-Edit, SEED-X-PPT, SEED-X-Story and SEED-X-Try-on. The proficient capabilities of these instruction-tuned model across various domains demonstrate the versatility of our pre-trained foundation model SEED-X. We perform instruction tuning on the foundation model SEED-X across different datasets, resulting in various models with distinct capabilities. Note that \textbf{we do not have an all-in-one instruction-tuned model that encompasses all abilities}, which will be explored for future work.

\section{Experiments}

\subsection{Quantitative Evaluation}
\noindent \textbf{Multimodal Comprehension.} We evaluate the multimodal comprehension capabilities of SEED-X-I on widely recognized image-based vision-language benchmarks, which include VQAv2 \cite{goyal2017vqav2}, GQA \cite{hudson2019gqa}, POPE \cite{li2023evaluating}, MME \cite{fu2023mme}, SEED \cite{li2023seed}, MMB \cite{liu2023mmbench}, MM-Vet \cite{yu2023mm}, and MMMU \cite{yue2024mmmu}. As listed in Tab.~\ref{tab:result_understanding}, SEED-X-I achieves competitive performance across various benchmarks, even when compared to MLLMs specifically designed for multimodal comprehension.

\noindent \textbf{Image Generation.} We evaluate the image generation capabilities of SEED-X-I on GenEval \cite{ghosh2024geneval}, which is a challenging benchmark to evaluate compositional image properties such as object co-occurrence, position, count, and color. As shown in Tab.~\ref{tab:geneval}, SEED-X obtains 51\% overall accuracy, demonstrating the model's excellent instruction-following capabilities for image generation.

\begin{table}[t]
    \centering
    \setlength{\tabcolsep}{4pt}
    \renewcommand{\arraystretch}{1.2}
    \scriptsize
    \caption{\textbf{Evaluation of text-to-image generation ability on GenEval benchmark}. ``Und.'' and ``Gen.'' denote ``understanding'' and ``generation'', respectively. 
    }
    \begin{tabular}{llcccccccc}
        \toprule
        \textbf{Type} & \textbf{Method}  & \textbf{Single Obj.} & \textbf{Two Obj.} & \textbf{Counting} & \textbf{Colors} & \textbf{Position} & \textbf{Color Attri.} & \textbf{Overall$\uparrow$} \\
        \midrule
        \multirow{8}{*}{\textit{Gen. Only}} 
        & LDM~\cite{rombach2022high}  & $0.92$ & $0.29$ & $0.23$ & $0.70$ & $0.02$ & $0.05$ & $0.37$ \\
        & SDv$1.5$~\cite{rombach2022high}  & $0.97$ & $0.38$ & $0.35$ & $0.76$ & $0.04$ & $0.06$ & $0.43$ \\
        & PixArt-$\alpha$~\cite{chen2023pixart}  & $0.98$ & $0.50$ & $0.44$ & $0.80$ & $0.08$ & $0.07$ & $0.48$ \\
        & SDv$2.1$~\cite{rombach2022high}  & $0.98$ & $0.51$ & $0.44$ & $0.85$ & $0.07$ & $0.17$ & $0.50$ \\
        & DALL-E $2$~\cite{ramesh2022hierarchical}  & $0.94$ & $0.66$ & $0.49$ & $0.77$ & $0.10$ & $0.19$ & $0.52$ \\
        & SDXL~\cite{podell2023sdxl}  & $0.98$ & $0.74$ & $0.39$ & $0.85$ & $0.15$ & $0.23$ & $0.55$ \\
        \midrule
        \multirow{2}{*}{\textit{Und. and Gen.}}
        & LWM~\cite{liu2024world}  & $0.93$ & $0.41$ & $0.46$ & $0.79$ & $0.09$ & $0.15$ & $0.47$ \\
        & \textbf{SEED-X}  & $0.96$ & $0.65$ & $0.31$ & $0.80$ & $0.18$ & $0.14$ & $0.51$ \\
        \bottomrule
    \end{tabular}
    \label{tab:geneval}
\end{table}

\begin{figure}[]
\vspace{-20pt}
	\centering
	\makebox[\textwidth][c]{\includegraphics[width=1.08\linewidth]{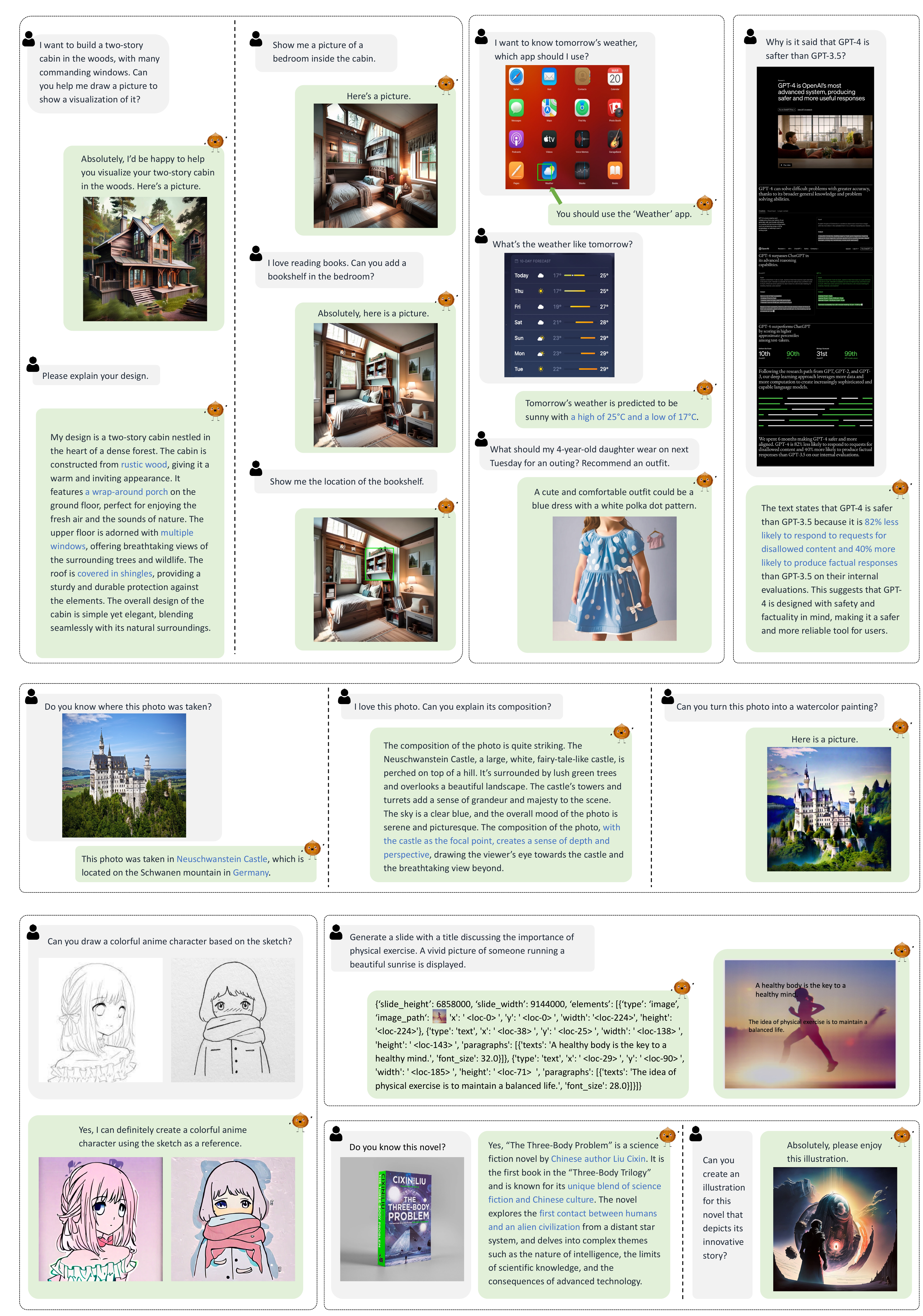}}%
\caption{Examples of what SEED-X can do in real-world scenarios after different instruction tuning through unifying multi-granularity comprehension and generation. Our instruction tuned models can function as an interactive designer, generating images without descriptive captions while
illustrating creative intent, and showcasing visualizations of modified images based on user's intent. They can act as knowledgeable personal assistants, comprehending images of arbitrary sizes and offering relevant suggestions in multi-turn conversations.}
	\label{fig:case}
\end{figure}

\begin{figure}[h!]
	\centering
	\includegraphics[width=1.0\linewidth]{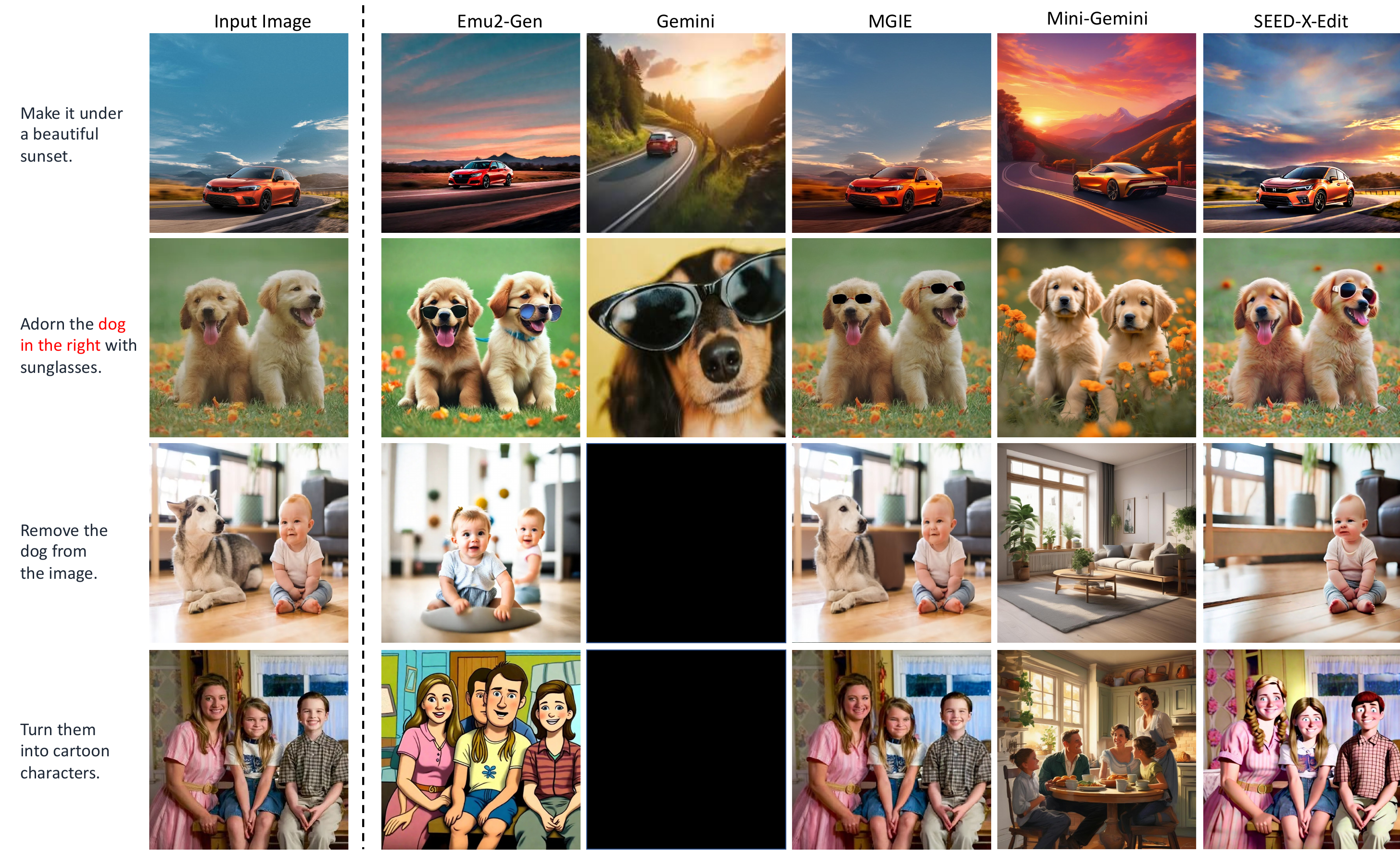}
\caption{Qualitative comparison between MLLMs for image manipulation. SEED-X-Edit shows enhanced ability in adhering to instructions while preserving low-level details of input images. The black images result from Gemini's inability to display human images.}
	\label{fig:edit}
\end{figure}

\subsection{Qualitative Evaluation}
\subsubsection{Applications in the Real World.}
SEED-X can be effectively instruction tuned to function as various multimodal AI assistants in the real world across different domains after integrating two enhanced features, including the comprehension of images of arbitrary sizes and ratios, and multi-granularity image generation, encompassing both high-level instructional image generation and low-level image manipulation tasks. As shown in Fig.~\ref{fig:teaser_example} and Fig.~\ref{fig:case}, our instruction tuned models can serve as an interactive designer, which can generate images without descriptive captions while illustrate creative intent, and showcase visualizations of modified images. For example, it can explain the design idea of concept image for AGI and a two-story cabin. It can create an imaginative illustration for the novel without the need of describing the scene with languages. It can further offer modification suggestions of the user's room and showcase the visualization. Additionally, the instruction tuned models can act as an knowledgeable personal assistant, comprehending images of arbitrary sizes and providing relevant suggestions. For example, it can identify foods suitable for fat reduction in the refrigerator, display appropriate clothing based on the screenshot of weather forecasts.

\subsubsection{Image Generation and Manipulation.} 
We compare previous MLLMs that are capable of generating images for text-to-image generation in Fig.~\ref{fig:gen} of Appendix. Our instruction tuned model can generate images that are more aligned with the elements in the caption and possess artistic qualities. Through utilizing a pre-trained ViT Tokenizer as the bridge to decouple the training of visual de-tokenizer and the MLLM, our pre-trained model SEED-X can effectively realize high-quality image generation, which is a fundamental capability to be applied in real-world scenarios.

 We compare image manipulation with previous MLLMs including Emu2-Gen~\cite{sun2023emu2}, Gemini~\cite{team2023gemini}, MGIE~\cite{team2023gemini} and Mini-Gemini~\cite{li2024mini}. As shown in Fig.~\ref{fig:edit}, we can observe that SEED-X-Edit can more effectively adhere to editing instructions while maintaining the low-level details of the input image. For instance, SEED-X-Edit can accurately add sunglasses to the dog on the right, while both Emu2-Gen and MGIE fail to follow the instruction, resulting in sunglasses being added to both dogs. Additionally, SEED-X-Edit successfully eliminates the dog in the baby image while preserving the background details and the baby's features. In contrast, Emu2-Gen fails to retain the fine details of the input image, and MGIE is unsuccessful in removing the dog. Note that Gemini lacks the ability to edit images as it \textbf{retrieves images} on the Internet. Here the presence of black images is due to its failure to display images related to human portraits. Mini-Gemini generates \textbf{text prompts} as the input of a pre-trained SDXL model, which can not preserve the visual details of the input image. The examples show the effectiveness of our instruction model for high-precision image manipulation. Our MLLM accurately predicts visual semantic representations based on an input image and a language instruction, which serve as input for the U-Net.  The visual de-tokenizer can further condition on the input image, ensuring the preservation of fine-grained details in the decoded images.

\subsubsection{Multimodal Comprehension.}  We provide qualitative examples of multimodal comprehension by SEED-X-I in Fig.~\ref{fig:com} and Fig.~\ref{fig:com2} of Appendix. SEED-X-I can realize fine-grained object detection and perception, text-rich comprehension, fundamental mathematical computation, world-knowledge and commonsense reasoning, diagram understanding, which are crucial capabilities for its application in real-world scenarios.

\begin{figure}[t]
	\centering
	\includegraphics[width=1.0\linewidth]{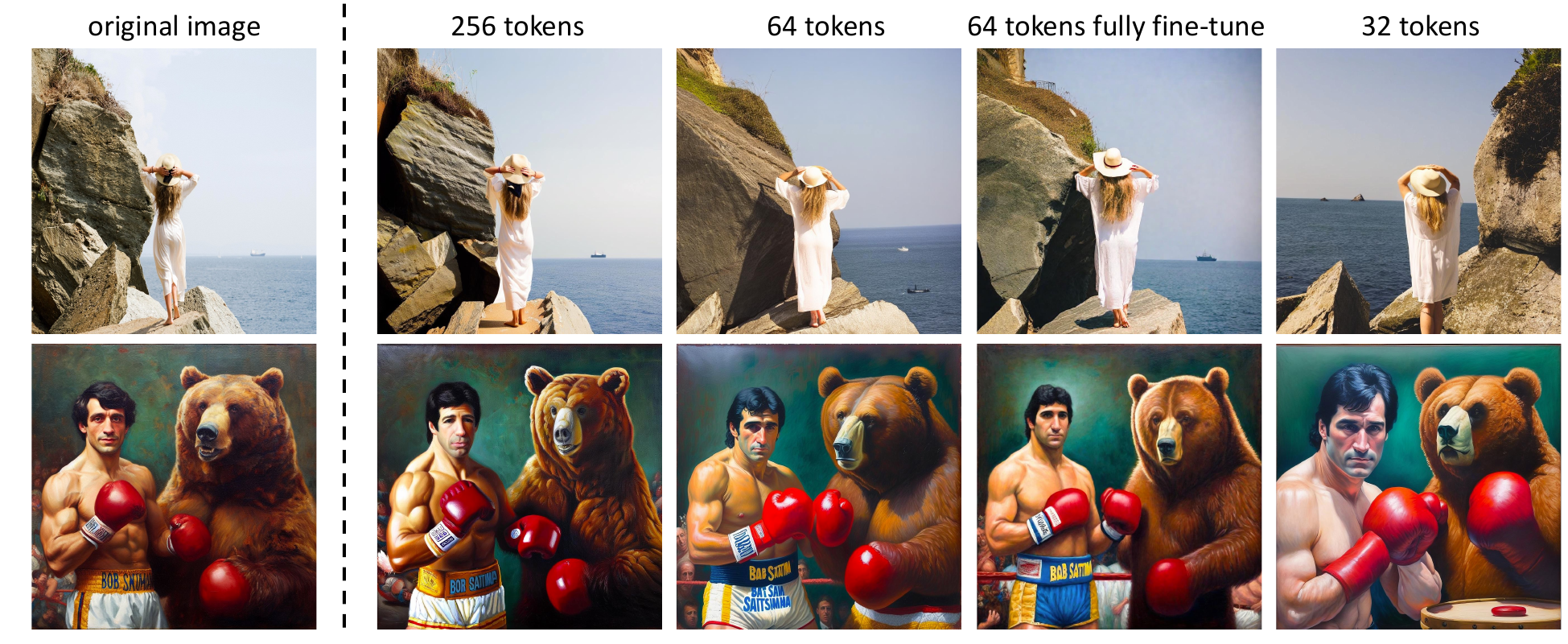}
   \vspace{-15pt}
\caption{Ablation study on the number of visual tokens and trainable parameters for training visual de-tokenizer.}
	\label{fig:ablation_detokenizer}
\end{figure}

\subsection{Ablation Study}\label{sec:ablation}
In this section, we perform ablation studies on the training of our visual de-tokenizer and the pre-training of SEED-X to enable a MLLM for image generation.
 
For visual de-tokenization, N visual embeddings (after average pooling) from the ViT tokenizer are fed into a learnable module as the inputs of the U-Net of the pre-trained SD-XL. We perform an ablation study on the number of visual tokens and the learnable parameters of the SD-XL U-Net, where keys and values within the U-Net are optimized if not specified with ``fully fine-tune''. As shown in Fig.~\ref{fig:ablation_detokenizer}, we can observe that more visual tokens can result in better reconstruction of the original images. For example, the decoded images from 256 visual embeddings can recover the characters' postures of the original images, while decoded images from 32 visual embeddings have already lost the original structure of the scene. We further observe that fully fine-tuning the parameters of the SD-XL U-Net can lead to distortions in image details, such as the woman's feet, compared to only training the keys and values within the U-Net. In SEED-X, we use N = 64 visual embeddings to train the visual de-tokenizer and only optimize the keys and values within the U-Net (See below for an explanation of why we do not choose N = 256).

\begin{figure}[t]
	\centering
	\includegraphics[width=1.0\linewidth]{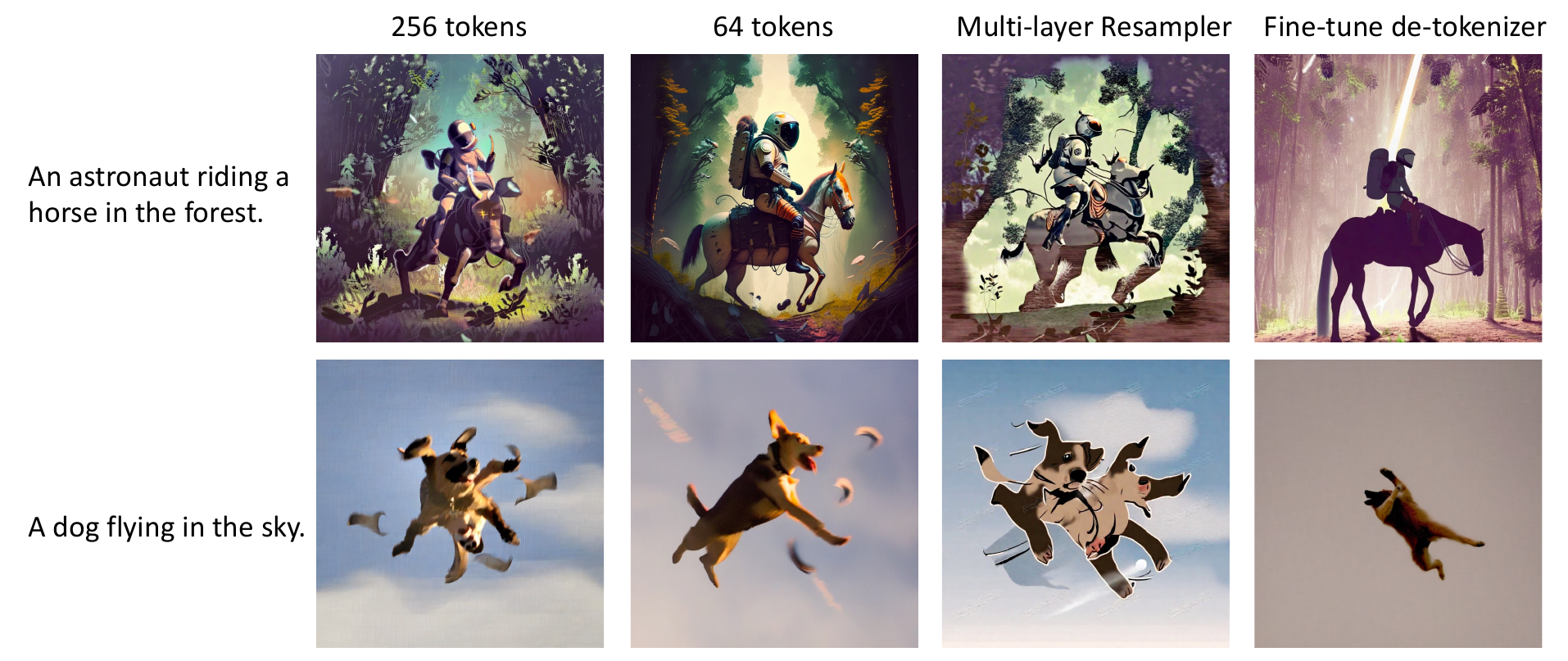}
  \vspace{-15pt}
\caption{Ablation study on the number of visual tokens, model architecture and optimization targets during pre-training SEED-X for image generation.}
	\label{fig:ablation_t2i}
\end{figure}

To enable MLLM for image generation, we employ N learnable queries to obtain the output visual representations from the LLM, which are trained to reconstruct N visual embeddings from the ViT tokenizer with a learnable module. We first perform an ablation study on the number of learnable queries. The images generated by the MLLM based on the input caption are shown in Fig.~\ref{fig:ablation_t2i}. We can observe that using 256 learnable queries to reconstruct 256 visual embeddings can lead to distortion in the generated images compared with N = 64. This occurs because regressing more visual features is more challenging for the model, even though 256 visual embeddings from the de-tokenizer can better reconstruct images, as demonstrated in the previous ablation study. We also observe that, compared to learning a one-layer cross-attention for reconstructing image features, a multi-layer resampler (multi-layer cross-attention) yields less satisfactory performance, which can happen due to the lack of more direct regularizations on the hidden states of the LLM. We further optimize the visual de-tokenizer by using the reconstructed visual embeddings from the MLLM as input instead of ViT features, but the generated images exhibit a more monotonous appearance. It demonstrates the effectiveness of utilizing the ViT Tokenizer as the bridge to decouple the training of visual de-tokenizer and the MLLM for image generation.

\section{Conclusion}
We present SEED-X, a versatile foundation model, which can serve as various multimodal AI assistants in the real world after instruction tuning. In order to make a multimodal foundation model applicable in open-world context, we integrate two enhanced features into SEED-X including image comprehension of arbitrary sizes and ratios, and multi-granularity image generation, which encompasses both high-level instructional image generation and low-level image manipulation. We hope that SEED-X can inspire future research into the potential of MLLMs in the real-world scenarios through unifying multi-granularity comprehension and generation.

{\small
\bibliographystyle{unsrt}
\bibliography{SEED-X}
}

\clearpage
\appendix

\section{Pre-training and Instruction Tuning Datasets}\label{sec:appendix_data}
As listed in Tab.~\ref{tab:dataset}, we pre-train SEED-X and conduct instruction tuning on a large variety of both public datasets and in-house data. For multimodal pre-training, we utilize image-caption pairs, grounded image-caption pairs, interleaved image and text content, OCR data and pure text data. The images of LAION-COCO~\cite{laion-coco} and SAM~\cite{kirillov2023segment} are re-captioned for a more detailed descriptive caption to improve both image comprehension and generation. 

For instruction tuning, we utilize various public VQA datasets, and further curate text-rich QA, grounded and referencing QA to enhance the model's capability of comprehending text-rich images and detecting objects that requires reasoning. We use multiple conversational datasets, which are specifically collected for MLLMs with open-form text output. We use the same image-caption pairs as in the pre-training phase to maintain the model's ability to generate images. For the image manipulation, since the high-precision editing dataset MagicBrush~\cite{zhang2023magicbrush} is only at the level of thousands, we employ a series of models to collect a dataset of millions of image editing examples, which are used for both training the visual de-tokenizer and SEED-X-Edit. We further collected data on slides, obtaining images, captions, and layouts for training slide generation.

\begin{table}[t]
\small
\centering
\setlength\tabcolsep{2pt}
\renewcommand{\arraystretch}{1.4} 
\caption{Overview of the pre-training and instruction tuning datasets. }
\label{tab:dataset}
\resizebox{0.95\columnwidth}{!}{
\begin{tabular}{l|l}
\toprule
Type                        & Dataset                                                                                                                                                   \\ \hline
\multicolumn{2}{l}{\textbf{Pre-training}} 
\\\hline
Image-Caption          & \begin{tabular}[c]{@{}l@{}}LAION-COCO~\cite{laion-coco} (Re-caption), SAM~\cite{kirillov2023segment} (Re-caption), Unsplash~\cite{unsplash}, \\ LAION-Aesthetics\cite{laion_aesthetics}, JourneyDB~\cite{pan2023journeydb}, CapFusion~\cite{yu2023capsfusion},  \end{tabular}   \\\hline
Grounded Image-Caption & GRIT~\cite{peng2023kosmos}                                                                                                                                                      \\\hline
Interleaved Image-Text  & \begin{tabular}[c]{@{}l@{}} MMC4~\cite{zhu2023multimodal}, OBELICS~\cite{laurencon2023obelics}, OpenFlamingo~\cite{awadalla2023openflamingo} \end{tabular}                                                                                                                                    \\\hline
OCR                         & LLaVAR~\cite{zhang2023llavar}, Slides (In-house)                                                                                                                                 \\\hline
Pure Text                   & Wikipedi                                                                                                                                                  \\\hline
\multicolumn{2}{l}{\textbf{Instruction Tuning}} \\\hline
VQA                         & \begin{tabular}[c]{@{}l@{}}LLaVAR~\cite{zhang2023llavar}, Text-rich QA (In-house), MIMIC-IT~\cite{li2023mimic}, MathQA~\cite{amini2019mathqa}, \\ChartQA~\cite{masry2022chartqa}, AI2D~\cite{kembhavi2016diagram}, ScienceQA~\cite{lu2022learn}, KVQA~\cite{shah2019kvqa}, \\DVQA~\cite{kafle2018dvqa}, Grounded QA (In-house), Referencing QA (In-house)\end{tabular} \\\hline
Conversation                & \begin{tabular}[c]{@{}l@{}}LLaVA-150k~\cite{liu2024visual}, ShareGPT~\cite{chen2023sharegpt4v}, VLIT~\cite{li2023vision},\\LVIS-Instruct4V~\cite{wang2023see}, Vision-Flan~\cite{xu2024vision}, ALLaVA-4V~\cite{chen2024allava}\end{tabular}                                           \\\hline
Image Generation          & \begin{tabular}[c]{@{}l@{}}LAION-COCO~\cite{laion-coco} (Re-caption), SAM~\cite{kirillov2023segment} (Re-caption), Unsplash~\cite{unsplash}, \\ LAION-Aesthetics\cite{laion_aesthetics}, JourneyDB~\cite{pan2023journeydb}  \end{tabular}                            \\\hline
Image Editing               & \begin{tabular}[c]{@{}l@{}}Instructpix2pix~\cite{brooks2023instructpix2pix}, MagicBrush~\cite{zhang2023magicbrush}, Openimages~\cite{kuznetsova2020open}-editing (In-house), \\Unsplash~\cite{unsplash}-editing (In-house)\end{tabular}                        \\\hline
Slides Generation                  & In-house data                                                                                                                                             \\\hline
Story Telling               & VIST~\cite{huang2016visual}                                                                                                                                                     \\\hline
Virtual Try-on               & VITON-HD~\cite{choi2021viton}     \\\hline                                                            \end{tabular}}
\end{table}

\section{Implementation Details}\label{sec:appendix_im}
\textbf{Visual Tokenization and De-tokenization.} We use the visual encoder from Qwen-vl~\cite{bai2023qwen} as the ViT Tokenizer and adopt 1D average pooling to obtain $N=64$ visual embeddings. These visual embeddings are fed into four layers of cross-attention as the input of the U-Net initialized from SDXL~\cite{podell2023sdxl}. In the first stage, we optimize the parameters of the cross-attention layers and the keys and values within the U-Net on the images from JourneyDB \cite{sun2024journeydb}, LAION-Aesthetics \cite{laion_aesthetics}, Unsplash \cite{unsplash}, and LAION-COCO \cite{laioncoco}. We train the visual de-tokenizer on 32 A100-40G GPUs with 42K training steps, where the learning rate is set to 1e-4 with cosine decay.

In the second stage, we encode the condition image into the latent space via the VAE
encoder, and concatenate them with the noisy latent as the input of U-Net. The channel
number of the U-Net convolutional layer is expanded from 4 to 8, and all parameters of
U-Net are optimized. We pre-train the visual conditioner on MagicBrush~\cite{zhang2023magicbrush} and in-house image editing data, as well as the image-caption pairs in the first stage, where the conditional inputs are set to zeros. We fine-tune the visual de-tokenizer on 32 A100-40G GPUs with 30K training steps, where the learning rate is set to 1e-4 with cosine decay.

\textbf{Multimodal Pre-training and Instruction Tuning.}
We utilize the visual encoder from Qwen-vl~\cite{bai2023qwen} as the ViT Tokenizer and initialize a cross-attention layer to obtain $N=64$ visual embedding as the input of the LLM initialized from Llama2-chat-13B. We initialize $N=64$ learnable queries and the output hidden states from them are fed into a cross-attention layer to reconstruct $N=64$ visual embeddings from the ViT Tokenizer. We optimize the LLM using LoRA and optimize the parameters of the input cross-attention layer, output cross-attention layer, extrapolatable 2D positional embeddings, and LoRA on image-captions pairs, grounded image-texts, interleaved image-text data,
OCR data and pure texts. We perform pre-training with 48 H800-80G GPUs (10 days) on a total of 158M samples, where the learning rate is set to 1e-4 with cosine decay.

For the instruction tuning, we fine-tune a LoRA module on the pre-trained model, and optimize the parameters of the input cross-attention layer, output cross-attention layer, extrapolatable 2D positional embeddings, and LoRA. We fine-tune SEED-X with conversational and image generation data to yield a general instruction-tuned model SEED-X-I. We further fine-tune SEED-X on specialized datasets, resulting in a series of instruction-tuned models tailored for specific tasks, including SEED-X-Edit, SEED-X-PPT, SEED-X-Story and SEED-X-Try-on.

\section{Qualitative Examples}\label{sec:qualitative}
\begin{figure}[t]
	\centering
	\includegraphics[width=1.0\linewidth]{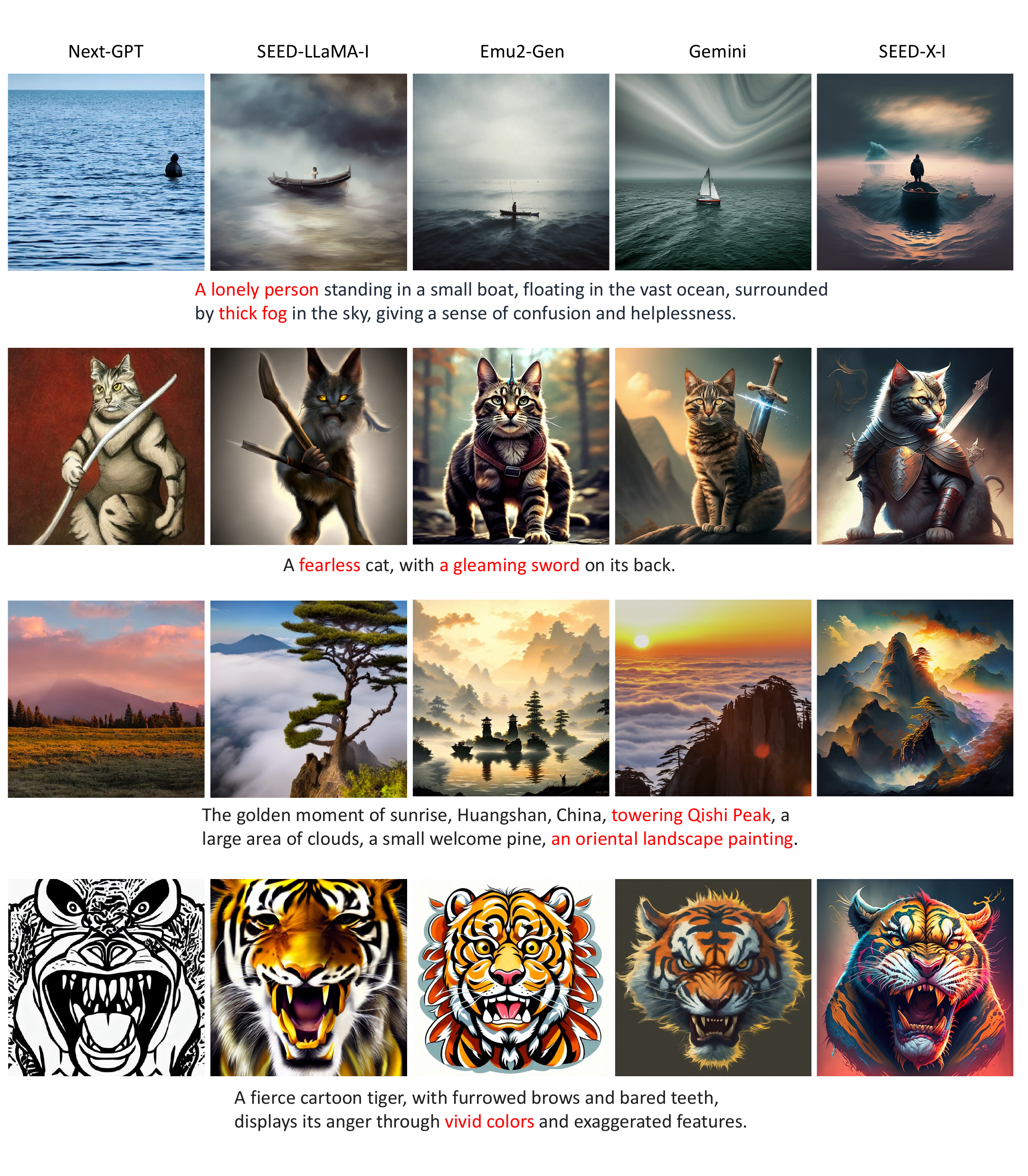}
   \vspace{-15pt}
\caption{Qualitative comparison between MLLMs for text-to-image generation. SEED-X-I is capable of generating images that are more closely aligned with the elements in the descriptive caption and exhibit artistic qualities.}
	\label{fig:gen}
\end{figure}

\textbf{Text-to-image Generation,} Fig.~\ref{fig:gen} visualizes the comparison between MLLMs for text-to-image generation including Next-GPT~\cite{wu2023nextgpt}, SEED-LLaMA-I\cite{ge2023making}, Emu2-Gen~\cite{sun2023emu2} and Gemini~\cite{team2023gemini}. Compared with previous MLLMs, our instruction tuned model can generate images that are more aligned with the elements in the descriptive caption and possess artistic qualities. For example, images generated by SEED-X-I vividly and accurately depicts ``person standing in a small boat'', ``a gleaming sword on its back'', ``an oriental landscape painting'', ``tiger with vivid colors'' in the captions. Through utilizing a pre-trained ViT Tokenizer as the bridge to decouple the training of visual de-tokenizer and the MLLM, our pre-trained model SEED-X can effectively realize high-quality image generation, which is a fundamental capability for applying multimodal models in real-world scenarios.

\textbf{Image Manipulation.} We compare image manipulation with previous MLLMs including Emu2-Gen~\cite{sun2023emu2}, Gemini~\cite{team2023gemini}, MGIE~\cite{team2023gemini} and Mini-Gemini~\cite{li2024mini}. Language-guided image manipulation presents a significant challenge as the model must be capable of comprehending free-form instructions and generating images with the low-level details of the input image preserved. As shown in Fig.~\ref{fig:edit}, we can observe that SEED-X-Edit can more effectively adhere to editing instructions while maintaining the low-level details of the input image. For instance, SEED-X-Edit can accurately add sunglasses to the dog on the right, while both Emu2-Gen and MGIE fail to follow the instruction, resulting in sunglasses being added to both dogs. Additionally, SEED-X-Edit successfully eliminates the dog in the baby image while preserving the low-level background details and the baby's features. In contrast, Emu2-Gen fails to retain the fine details of the input image, and MGIE is unsuccessful in removing the dog. Note that Gemini lacks the ability to edit images as it \textbf{retrieves images} on the Internet. Here the presence of black images is due to its failure to display images related to human portraits. Mini-Gemini generates \textbf{text prompts} as the input of a pre-trained SDXL model, which can not preserve the visual details of the input image. The examples demonstrate the effectiveness of our instruction model for high-precision image manipulation. Our MLLM accurately predicts visual semantic representations based on an input image and a language instruction, which serve as input for the U-Net.  The visual de-tokenizer can further condition on the input image, ensuring the preservation of fine-grained details in the decoded images. 

\begin{figure}[t]
	\centering
	\includegraphics[width=1.0\linewidth]{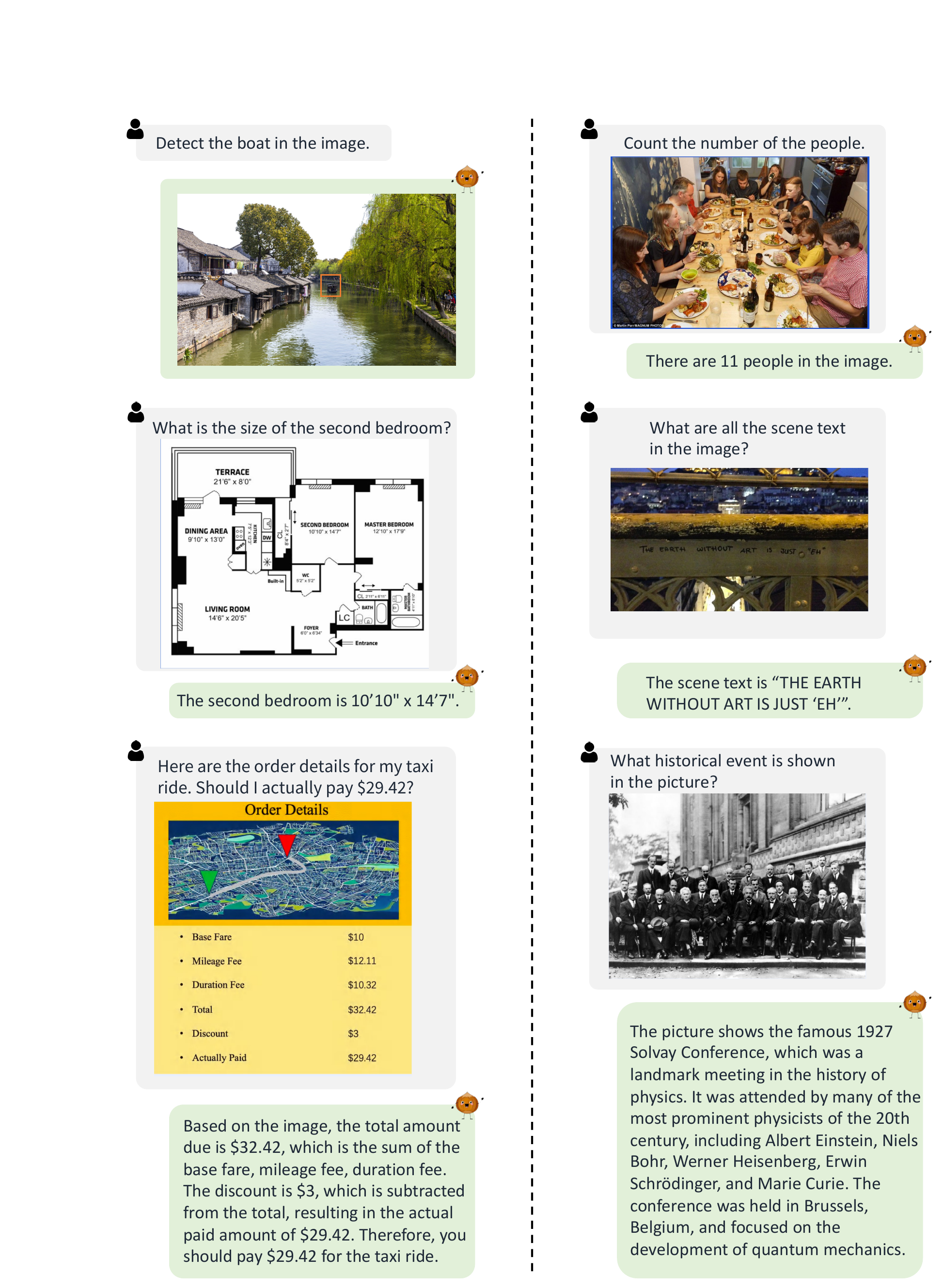}
\caption{Qualitative examples of multimodal comprehension by SEED-X-I.}
	\label{fig:com}
\end{figure}

\begin{figure}[t]
	\centering
	\includegraphics[width=1.0\linewidth]{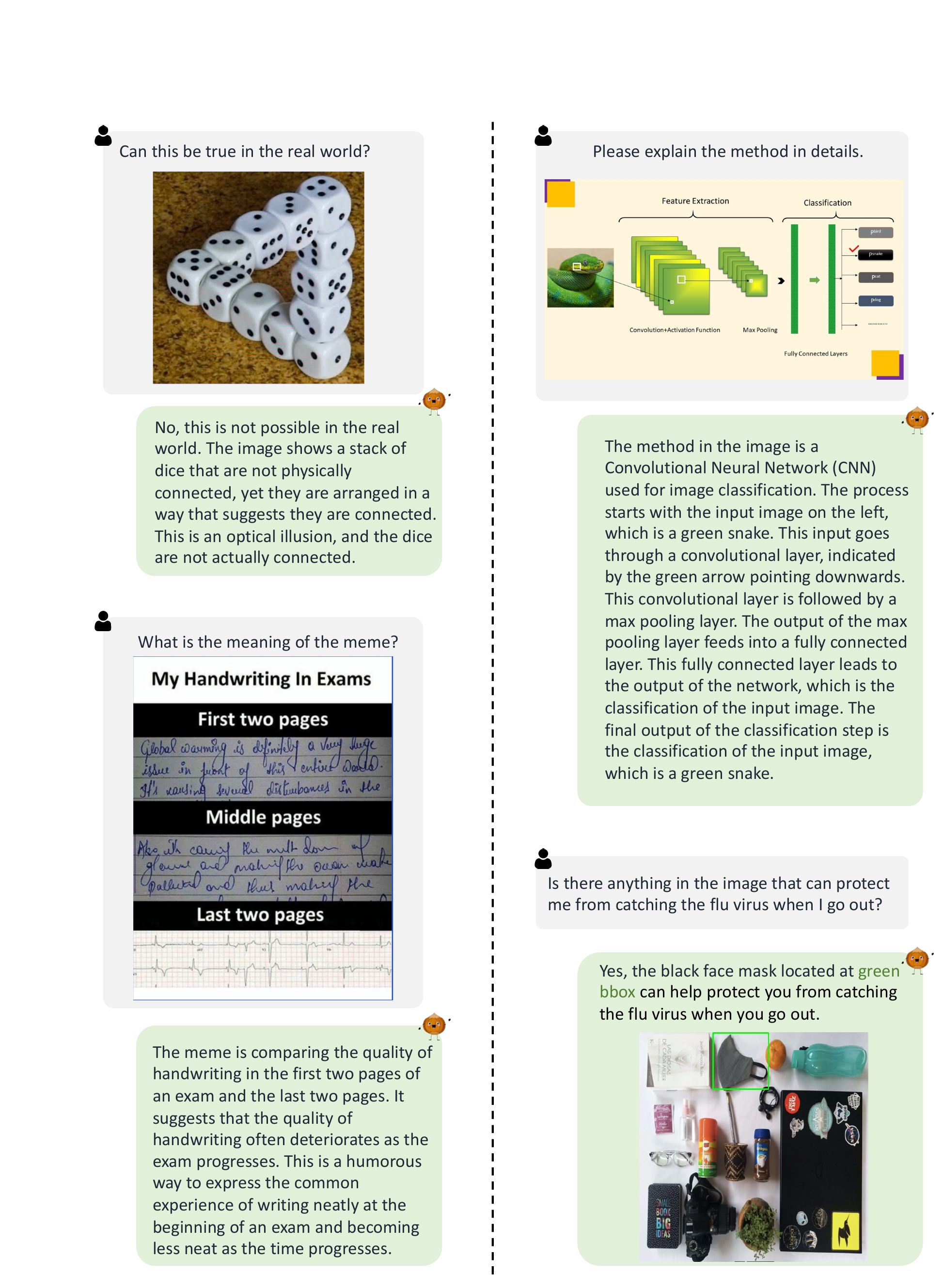}
\caption{Qualitative examples of multimodal comprehension by SEED-X-I.}
	\label{fig:com2}
\end{figure}

\textbf{Multimodal Comprehension}  We show qualitative examples of multimodal comprehension by SEED-X-I in Fig.~\ref{fig:com} and Fig.~\ref{fig:com2}. SEED-X-I can realize fine-grained object detection and perception, text-rich comprehension, fundamental mathematical computation, world-knowledge and commonsense reasoning, diagram understanding, which are crucial capabilities for its application in the real world.

\end{document}